\pdfoutput=1
\documentclass[11pt,a4paper]{article}
\usepackage[hyperref]{acl2019}
\usepackage{times}
\usepackage{latexsym}
\usepackage{amsmath}
\usepackage[linesnumbered, ruled]{algorithm2e}
\usepackage{xcolor}

\usepackage{url}
\usepackage{multicol}
\usepackage{multirow}
\usepackage{booktabs}
\usepackage{tikz}
\usetikzlibrary{shapes.geometric, arrows}
\usetikzlibrary{positioning}
\usepackage{subfig}
\newenvironment{itemize*}%
  {\begin{itemize}%
    \setlength{\itemsep}{0pt}%
    \setlength{\parskip}{0pt}}%
  {\end{itemize}}
\newenvironment{enumerate*}%
  {\begin{enumerate}%
    \setlength{\itemsep}{2pt}%
    \setlength{\parskip}{2pt}}%
  {\end{enumerate}}

\newenvironment{enumerate**}%
  {\begin{enumerate}%
    \setlength{\itemsep}{0pt}%
    \setlength{\parskip}{0pt}}%
  {\end{enumerate}}
\aclfinalcopy 

\title{Style Transformer: Unpaired Text Style Transfer without \\Disentangled Latent Representation}

\author{Ning Dai, Jianze Liang, Xipeng Qiu\thanks{\ \  Corresponding author} , Xuanjing Huang \\
  Shanghai Key Laboratory of Intelligent Information Processing, Fudan University \\
  School of Computer Science, Fudan University \\
  825 Zhangheng Road, Shanghai, China \\
  \texttt{\{ndai16,jzliang18,xpqiu,xjhuang\}@fudan.edu.cn} }

\date{}

\begin{document}
\maketitle
\begin{abstract}
Disentangling the content and style in the latent space is prevalent in unpaired text style transfer. However, two major issues exist in most of the current neural models. 1) It is difficult to completely strip the style information from the semantics for a sentence. 2) The recurrent neural network (RNN) based encoder and decoder, mediated by the latent representation, cannot well deal with the issue of the long-term dependency, resulting in poor preservation of non-stylistic semantic content.
In this paper, we propose the Style Transformer, which makes no assumption about the latent representation of source sentence and equips the power of attention mechanism in Transformer to achieve better style transfer and better content preservation. Source code will be available on Github\footnote{\url{https://github.com/fastnlp/style-transformer}}.
\end{abstract}


\section{Introduction}

Text style transfer is the task of changing the stylistic properties (e.g.,
sentiment) of the
text while retaining the style-independent content
within the context. Since the definition of the text style is vague, it is difficult to construct paired sentences with the same content and differing styles. Therefore, the studies of text style transfer focus on the unpaired transfer.

Recently, neural networks have become the dominant methods in text style transfer. Most of the previous methods
\cite{hu2017toward,shen2017style,Fu18,
DBLP:journals/corr/abs-1711-04731,
DBLP:journals/corr/abs-1808-07894,
DBLP:conf/naacl/ZhangDS18,
DBLP:conf/acl/TsvetkovBSP18,
2019arXiv190111333J,
DBLP:journals/corr/abs-1711-09395,
DBLP:conf/acl/SantosMP18} formulate the style transfer problem into the ``encoder-decoder'' framework. The encoder maps the text into a style-independent latent representation (vector representation), and the decoder generates a new text with the same content but a different style from the disentangled latent representation plus a style variable.

These methods focus on how to disentangle the content and style in the latent space. The latent representation needs better preserve the meaning of the text while reducing its stylistic properties.
Due to lacking paired sentence, an adversarial loss \cite{goodfellow2014generative} is used in the latent
space to discourage encoding style information in the latent representation.
Although the disentangled latent representation brings better interpretability, in this paper, we address the following concerns for these models.

1) It is difficult to judge the quality of disentanglement. As reported in \cite{DBLP:conf/emnlp/ElazarG18, lample2018multipleattribute}, the style information can be still recovered from the latent representation even the model has trained adversarially. Therefore, it is not easy to disentangle the stylistic property from the semantics of a sentence.

2) Disentanglement is also unnecessary. \citet{lample2018multipleattribute} reported that a good decoder can generate the text with the desired style from an entangled latent representation by ``overwriting'' the original style.

3) Due to the limited capacity of vector representation, the latent representation is hard to capture the rich semantic information, especially for the long text. The recent progress of neural machine translation also proves that it is hard to recover the target sentence from the latent representation without referring to the original sentence.

4) To disentangle the content and style information in the latent space, all of the existing approaches have to assume the input sentence is encoded by a fix-sized latent vector. As a result, these approaches can not directly apply the attention mechanism to enhance the ability to preserve the information in the input sentence.

5) Most of these models adopt recurrent neural networks (RNNs) as encoder and decoder, which has a weak ability to capture the long-range dependencies between words in a sentence. Besides, without referring the original text, RNN-based decoder is also hard to preserve the content. The generation quality for long text is also uncontrollable.


In this paper, we address the above concerns of disentangled models for style transfer. Different from them, we propose Style Transformer, which takes
Transformer \cite{DBLP:conf/nips/VaswaniSPUJGKP17} as the basic block. Transformer is a fully-connected self-attention neural architecture, which has achieved many exciting results on natural language processing (NLP) tasks, such as machine translation \cite{DBLP:conf/nips/VaswaniSPUJGKP17}, language modeling \cite{DBLP:journals/corr/abs-1901-02860}, text classification \cite{DBLP:journals/corr/abs-1810-04805}.
Different from RNNs, Transformer uses stacked self-attention and point-wise, fully connected layers for both the encoder and decoder. Moreover, Transformer decoder fetches the information from the encoder part via attention mechanism, compared to a fixed size vector used by RNNs. With the strong ability of Transformer, our model can transfer the style of a sentence while better preserving its meaning.
The difference between our model and the previous model is shown in Figure \ref{fig:arch}.

Our contributions are summarized as follows:
\begin{itemize*}
    \item We introduce a novel training algorithm which makes no assumptions about the disentangled latent representations of the input sentences, and thus the model can employ attention mechanisms to improve its performance further.
    \item To the best of our knowledge, this is the first work that applies the Transformer architecture to style transfer task.
   \item Experimental results show that our proposed approach generally outperforms the other approaches on two style transfer datasets. Specifically, to the content preservation, Style Transformer achieves the best performance with a significant improvement.
\end{itemize*}

\section{Related Work}

Recently, many text style transfer approaches have been proposed. Among these approaches, there is a line of works aims to infer a latent representation for the input sentence, and manipulate the style of the generated sentence based on this learned latent representation. \citet{shen2017style} propose a cross-aligned auto-encoder
with adversarial training to learn a shared
latent content distribution and a separated latent style distribution. \citet{hu2017toward} propose a new neural generative model which combines variational
auto-encoders and holistic attribute discriminators for the effective imposition of semantic structures.
Following their work, many methods \cite{Fu18,john2018disentangled,DBLP:conf/naacl/ZhangDS18,DBLP:journals/corr/abs-1808-07894} has been proposed
based on standard encoder-decoder architecture. 

Although, learning a latent representation will make the model more interpretable and easy to manipulate, the model which is assumed a fixed size latent representation cannot utilize the information from the source sentence anymore.

On the other hand, there are also some approaches without manipulating latent representation are proposed recently. \citet{xu2018unpaired} propose a cycled reinforcement learning method
for unpaired sentiment-to-sentiment translation task. 
\citet{li2018delete} propose a three-stage method. Their model first extracts
content words by deleting phrases a strong attribute value, then retrieves new phrases associated with the target
attribute, and finally uses a neural model to
combine these into a final output. \citet{lample2018multipleattribute} reduce text style transfer to unsupervised machine translation problem \cite{DBLP:conf/emnlp/LampleOCDR18}. They employ Denoising Auto-encoders \cite{DBLP:conf/icml/VincentLBM08} and back-translation \cite{DBLP:conf/acl/SennrichHB16} to build a translation style between different styles.

 However, both lines of the previous models make few attempts to utilize the attention mechanism to refer the long-term history or the source sentence, except \citet{lample2018multipleattribute}. In many NLP tasks, especially for text generation, attention mechanism has been proved to be an essential technique to enable the model to capture the long-term dependency \cite{DBLP:journals/corr/BahdanauCB14, DBLP:conf/emnlp/LuongPM15, DBLP:conf/nips/VaswaniSPUJGKP17}.

In this paper, we follow the second line of work and propose a novel method which makes no assumption about the latent representation of source sentence and takes the proven self-attention network, Transformer, as a basic module to train a style transfer system.

\tikzstyle{transformer} = [rectangle, minimum width=3cm, minimum height=0.6cm, text centered, draw=black, fill=green!20]
\tikzstyle{nn} = [rectangle, minimum width=1cm, minimum height=0.6cm, text centered, draw=black, fill=green!20]

\tikzstyle{discriminator} = [rectangle, rounded corners, minimum width=1.2cm, minimum height=1.2cm,text centered, draw=black, fill=orange!30]
\tikzstyle{input} = [rectangle, minimum width=0.5cm, minimum height=0.5cm, text centered, draw=black, fill=yellow!30]
\tikzstyle{output} = [rectangle, rounded corners, minimum width=0.5cm, minimum height=0.5cm, text centered, draw=black, fill=blue!20]
\tikzstyle{latent} = [rectangle, minimum width=0.5cm, minimum height=0.5cm, text centered, draw=black, fill=gray!20]
\tikzstyle{style} = [rectangle, minimum width=0.5cm, minimum height=0.5cm, text centered, draw=black, fill=blue!20]
\tikzstyle{arrow} = [thick,->,>=stealth]
\tikzstyle{d_arrow} = [dashed,thick,->,>=stealth]

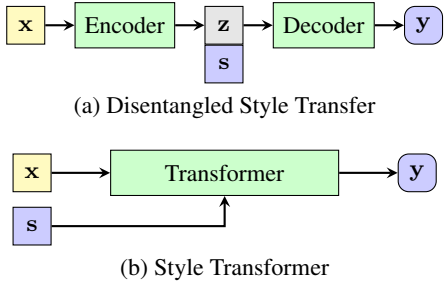
\begin{figure}[t]
    \centering
    \subfloat[Disentangled Style Transfer]{
    \begin{tikzpicture}[node distance=1em,font=\small]
    \node (input1) [input] {$\mathbf{x}$} ;
    \node (enc) [nn, right =of input1] {Encoder};

    \node (latent) [latent, right =of enc] {$\mathbf{z}$};

    \node (dec) [nn, right =of latent] {Decoder};

    \node (output1) [output,fill=blue!20, right =of dec] {$\mathbf{y}$};

    \node (style1) [style, node distance=0em,below =of latent] {$\mathbf{s}$} ;
    \draw [arrow] (input1) -- (enc);
    \draw [arrow] (enc) -- (latent);
    \draw [arrow] (latent) -- (dec);
    \draw [arrow] (dec) -- (output1);

    \end{tikzpicture}
    }\\
   \subfloat[Style Transformer]{
    \begin{tikzpicture}[node distance=2em,font=\small]
    \node (input1) [input] {$\mathbf{x}$} ;
    \node (trans1) [transformer, right =of input1] {Transformer} ;
    \node (output1) [output,fill=blue!20, right =of trans1] {$\mathbf{y}$};

    \node (style1) [style, node distance=0.5em,below =of input1,] {$\mathbf{s}$} ;
    \draw [arrow] (input1) -- (trans1);
    \draw [arrow] (trans1) -- (output1);
    \draw [arrow] (style1) -| (trans1);

    \end{tikzpicture}
    }
    \caption{General illustration of previous models and our model. $\mathbf{z}$ denotes style-independent content vector and $\mathbf{s}$ denotes the style variable.}
    \label{fig:arch}
\end{figure}

\section{Style Transformer}
To make our discussion more clearly, in this section, we will first give a brief introduction to the style transfer task, and then start to discuss our proposed model based on our problem definition.

\subsection{Problem Formalization}
In this paper, we define the style transfer problem as follows:
Considering a bunch of datasets $\{\mathcal{D}_{i}\}_{i=1}^{K}$, and each dataset $\mathcal{D}_{i}$ is composed of many natural language sentences. For all of the sentences in a single dataset $\mathcal{D}_{i}$ , they share some specific characteristic (e.g. they are all the positive reviews for a specific product), and we refer this shared characteristic as the \emph{style} of these sentences. In other words, a style is defined by the distribution of a dataset. Suppose we have $K$ different datasets $\mathcal{D}_{i}$, then we can define $K$ different styles, and we denote each style by the symbol $\mathbf{s}^{(i)}$. The goal of style transfer is that: given a arbitrary natural language sentence $\mathbf{x}$ and a desired style $\mathbf{\widehat{s}} \in \{\mathbf{s}^{(i)}\}_{i=1}^{K}$, rewrite this sentence to a new one $\widehat{\mathbf{x}}$ which
has the style $\mathbf{\widehat{s}}$ and preserve the information in original sentence $\mathbf{x}$ as much as possible.

\subsection{Model Overview}
To tackle the style transfer problem we defined above, our goal is to learn a mapping function $f_{\theta}(\mathbf{x}, \mathbf{s})$ where $\mathbf{x}$ is a natural language sentence and $\mathbf{s}$ is a style control variable. The output of this function is the transferred sentence $\widehat{\mathbf{x}}$ for the input sentence $\mathbf{x}$.

A big challenge in the text style transfer is that we have no access to the parallel corpora. Thus we can't directly obtain supervision to train our transfer model. In section \ref{sec:discriminator}, we employ two discriminator-based approaches to create supervision from non-parallel corpora.

Finally, we will combine the Style Transformer network and discriminator network via an overall learning algorithm in section \ref{sec:learning_algo} to train our style transfer system.

\subsection{Style Transformer Network}
\label{sec:Transformer_net}
Generally, Transformer follows the standard encoder-decoder architecture. Explicitly, for a input sentence $\mathbf{x} = (x_1, x_2, ..., x_n)$, the Transformer encoder $Enc(\mathbf{x}; \theta_{E})$ maps inputs to a sequence of continuous representations $\mathbf{z} = (z_1, z_2, ..., z_n)$. And the Transformer decoder $Dec(\mathbf{z}; \theta_{D})$ estimates the conditional probability for the output sentence $\mathbf{y} = (y_1, y_2, ..., y_n)$ by auto-regressively factorized its as:
\begin{equation}
p_{\theta}(\mathbf{y} | \mathbf{x}) = \prod_{t=1}^{m} p_{\theta}(y_t |  \mathbf{z}, y_1, ..., y_{t-1}).
\end{equation}

At each time step $t$, the probability of the next token is computed by a softmax classifier:
\begin{equation}
\label{eq:softmax}
p_{\theta}(y_t |  \mathbf{z}, y_1, ..., y_{t-1}) = \text{softmax}(\mathbf{o}_t),
\end{equation}
where $\mathbf{o}_t$ is logit vector outputted by decoder network.

To enable style control in the standard Transformer framework, we add a extra style embedding as input to the Transformer encoder $Enc(\mathbf{x}, \mathbf{s}; \theta_{E})$. Therefore the network can compute the probability of the output condition both on the input sentence $\mathbf{x}$ and the style control variable $\mathbf{s}$. Formally, this can be expressed as:
\begin{equation}
p_{\theta}(\mathbf{y} | \mathbf{x}, \mathbf{s}) = \prod_{t=1}^{m} p_{\theta}(y_t |  \mathbf{z}, y_1, ..., y_{t-1}),
\end{equation}
and we denote the predicted output sentence of this network by $f_{\theta}(\mathbf{x}, \mathbf{s})$.

\subsection{Discriminator Network}
\label{sec:discriminator}
Suppose we use $\mathbf{x}$ and $\mathbf{s}$ to denote the sentence and its style from the dataset $\mathcal{D}$. Because of the absence of the parallel corpora, we can't directly obtain the  supervision for the case $f_{\theta}(\mathbf{x}, \mathbf{\widehat{s}})$ where $\mathbf{s} \neq \mathbf{\widehat{s}}$. Therefore, we introduce a discriminator network to learn this supervision from the non-parallel copora.

The intuition behind the training of discriminator is based on the assumption below: As we mentioned above, we only have the supervision for the case $f_{\theta}(\mathbf{x}, \mathbf{s})$. In this case, because of the input sentence $\mathbf{x}$ and chosen style $\mathbf{s}$ are both come from the same dataset $\mathcal{D}$, one of the optimum solutions, in this case, is to reproduce the input sentence. Thus, we can train our network to reconstruct the input in this case. In the case of $f_{\theta}(\mathbf{x}, \mathbf{s})$ where $\mathbf{s} \neq \mathbf{\widehat{s}}$, we construct supervision from two ways. 1) For the content preservation, we train the network to reconstruct original input sentence $\mathbf{x}$ when we feed transferred sentence $\mathbf{\widehat{y}} = f_{\theta}(\mathbf{x}, \mathbf{\widehat{s}})$ to the Style Transformer network with the original style label $\mathbf{s}$. 2) For the style controlling, we train a discriminator network to assist the Style Transformer network to better control the style of the generated sentence.

In short, the discriminator network is another Transformer encoder, which learns to  distinguish the style of different sentences. And the Style Transformer network receives style supervision from this discriminator. To achieve this goal, we experiment with two different discriminator architectures.

\paragraph{Conditional Discriminator} In a setting similar to Conditional GANs \cite{cond_gan}, discriminator makes decision condition on a input style. Explicitly, a sentence $\mathbf{x}$ and a proposal style $\mathbf{s}$ are feed into discriminator $d_{\phi}(\mathbf{x}, \mathbf{s} )$, and the discriminator  is asked to answer whether the input sentence has the corresponding style. In discriminator training stage, the real sentence from datasets $\mathbf{x}$, and the reconstructed sentence $\mathbf{y} = f_{\theta}(\mathbf{x}, \mathbf{s})$ are labeled as \emph{positive}, and the transferred sentences $\mathbf{\widehat{y}} = f_{\theta}(\mathbf{x}, \mathbf{\widehat{s}})$ where $\mathbf{s} \neq \mathbf{\widehat{s}}$, are labeled as \emph{negative}. In Style Transformer network training stage, the network $f_{\theta}$ is trained to maximize the probability of \emph{positive} when feed $f_{\theta}(\mathbf{x}, \mathbf{\widehat{s}})$ and $\mathbf{\widehat{s}}$ to the discriminator.

\paragraph{Multi-class Discriminator}
Different from the previous one, in this case, only one sentence is feed into  discriminator $d_{\phi}(\mathbf{x})$, and the discriminator aims to answer the style of this sentence. More concretely, the discriminator is a classifier with  $K + 1$ classes. The first $K$ classes represent $K$ different styles, and the last class is stand for the generated data from $f_{\theta}(\mathbf{x}, \mathbf{\widehat{s}})$ , which is also often referred as fake sample. In discriminator training stage, we label the real sentences $\mathbf{x}$ and reconstructed sentences $\mathbf{y} = f_{\theta}(\mathbf{x}, \mathbf{s})$ to the label of the corresponding  style. And for the transferred sentence  $\mathbf{\widehat{y}} = f_{\theta}(\mathbf{x}, \mathbf{\widehat{s}})$ where $\mathbf{s} \neq \mathbf{\widehat{s}}$, is labeled as the class $0$. In Style Transformer network learning stage, we train the network $f_{\theta}(\mathbf{x}, \mathbf{\widehat{s}})$ to maximize the probability of the class which is stand for style $ \mathbf{\widehat{s}}$.

\tikzstyle{transformer} = [rectangle, minimum width=3cm, minimum height=0.7cm, text centered, draw=black, fill=green!20]
\tikzstyle{discriminator} = [rectangle, rounded corners, minimum width=1.2cm, minimum height=1cm,text centered, draw=black, fill=orange!30]
\tikzstyle{input} = [rectangle, minimum width=0.5cm, minimum height=0.5cm, text centered, draw=black, fill=yellow!30]
\tikzstyle{output} = [rectangle, rounded corners, minimum width=0.5cm, minimum height=0.5cm, text centered, draw=black, fill=blue!20]
\tikzstyle{style} = [rectangle, minimum width=0.5cm, minimum height=0.5cm, text centered, draw=black, fill=yellow!30]
\tikzstyle{arrow} = [thick,->,>=stealth]
\tikzstyle{d_arrow} = [dashed,thick,->,>=stealth]

\begin{figure}
    \centering
    \begin{tikzpicture}[node distance=2.3cm]
    \node (input1) [input] {$\mathbf{x}$} ;
    \node (trans1) [transformer, right of=input1] {$f_{\theta}(\mathbf{x},\mathbf{s})$} ;
    \node (output1) [output,fill=yellow!30, right of=trans1] {$\mathbf{y}$};

    \node (style1) [style, below of=trans1, yshift=+1.4cm] {$\mathbf{s}$} ;

    \node (trans3) [transformer, below of=style1, yshift=+1.4cm] {$f_{\theta}(\widehat{\mathbf{y}},\mathbf{s})$} ;
    \node (output3) [output, , fill=yellow!30, left of=trans3] {$\mathbf{y}$};

    \node (trans2) [transformer, below of=trans3, node distance=3em] {$f_{\theta}(\mathbf{x},\widehat{\mathbf{s}})$} ;
    \node (input2) [input, left of=trans2] {$\mathbf{x}$} ;
    \node (output2) [output, right of=trans2, yshift=+0.6cm] {$\widehat{\mathbf{y}}$};

    \node (style2) [style, fill=blue!20,below of=trans2, node distance=3em] {$\widehat{\mathbf{s}}$} ;
    \node (disc) [discriminator, right of=style2, xshift=-0.4cm] {$d_{\phi}(\widehat{\mathbf{y}})$};

    \draw [arrow] (input1) -- (trans1);
    \draw [arrow] (trans1) -- (output1);
    \draw [d_arrow] (output1) -- ++(0,+0.7) -- node[anchor=south] {$\mathcal{L}_{self}$}(0,+0.7)  -- (input1);

    \draw [arrow] (style1) -- (trans1);
    \draw [arrow] (style1) -- (trans3);

    \draw [arrow] (input2) -- (trans2);
    \draw [arrow] (trans2) -| (output2);
    \draw [arrow] (output2) |- (trans3);
    \draw [arrow] (output2) -- ++(+0.5,0) |- (disc.east);
    \draw [d_arrow] (disc) -- node[anchor=north] {$\mathcal{L}_{style}$}(style2);
    \draw [arrow] (style2) -- (trans2);
    \draw [arrow] (trans3) -- (output3);
    \draw [d_arrow] (output3) -- node[anchor=east] {$\mathcal{L}_{cycle}$}(input2);

    \end{tikzpicture}
    \caption{The training process for Style Transformer network. The input sentence $\mathbf{x}$ and input style $\mathbf{s}(\mathbf{\widehat{s}})$ is feed into Transformer network $f_{\theta}$. If the input style $\mathbf{s}$ is the same as the style of sentence $\mathbf{x}$, generated sentence $\mathbf{y}$ will be trained to reconstruct $\mathbf{x}$. Otherwise, the generated sentence $\mathbf{\widehat{y}}$ will be feed into Transformer $f_{\theta}$ and discriminator $d_{\phi}$ to reconstruct input sentence $\mathbf{x}$ and input style $\mathbf{\widehat{s}}$ respectively.}
    \label{fig:lean}
\end{figure}
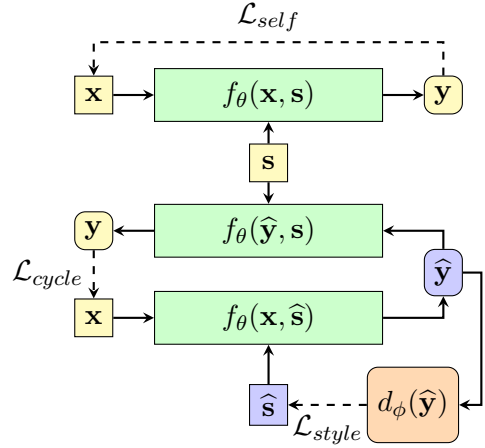

\subsection{Learning Algorithm}
\label{sec:learning_algo}
In this section, we will discuss how to train these two networks. And the training algorithm of our model can be divided into two parts: the discriminator learning and Style Transformer network learning. The brief illustration is shown in Figure \ref{fig:lean}.

\subsubsection{Discriminator Learning}
Loosely speaking, in the discriminator training stage, we train our discriminator to distinguish between the real sentence $\mathbf{x}$ and reconstructed sentence $\mathbf{y} = f_{\theta}(\mathbf{x}, \mathbf{s})$ from the transferred sentence $\mathbf{\widehat{y}} = f_{\theta}(\mathbf{x},\mathbf{\widehat{s}})$. The loss function for the discriminator is simply the cross-entropy loss of the classification problem.

For the conditional discriminator:
\begin{equation}
\label{eq:cond_cls_loss}
\mathcal{L}_{discriminator}(\phi) = - p_{\phi}(\mathbf{c} | \mathbf{x}, \mathbf{s}).
\end{equation}

And for the multi-class discriminator:
\begin{equation}
\label{eq:multi_cls_loss}
\mathcal{L}_{discriminator}(\phi) = - p_{\phi}(\mathbf{c} | \mathbf{x}).
\end{equation}

According to the difference of discriminator architecture, there is a different protocol for how to label these sentences, and the details can be found in Algorithm \ref{alg:discriminator}.

\begin{algorithm}\small
\caption{Discriminator Learning}
\label{alg:discriminator}
\LinesNumbered
\KwIn{Style Transformer $f_{\theta}$, discriminator $d_{\phi}$, and a dataset $\mathcal{D}_i$ with style $\mathbf{s}$}

Sample a minibatch of m sentences $\{\mathbf{x}_1, \mathbf{x}_2, ...\mathbf{x}_m\}$ from $\mathcal{D}_i$. \;
\ForEach{$\mathbf{x} \in \{\mathbf{x}_1, \mathbf{x}_2, ...\mathbf{x}_m\}$}{
    Randomly sample a style $\mathbf{\widehat{s}} (\mathbf{s} \neq \mathbf{\widehat{s}})$\;
    Use $f_{\theta}$ to generate two new sentence\\ $\mathbf{y} = f_{\theta}(\mathbf{x}, \mathbf{s})$\\
    $\mathbf{\widehat{y}} = f_{\theta}(\mathbf{x}, \mathbf{\widehat{s}})$ \;
    \eIf{$d_{\phi}$ is conditional discriminator}
    {
        Label $\{(\mathbf{x}, \mathbf{s}), (\mathbf{y}, \mathbf{s})\}$ as 1 \;
        Label $\{(\mathbf{x}, \mathbf{\widehat{s}}), (\mathbf{\widehat{y}}, \mathbf{\widehat{s}})\}$ as 0 \;
    }
    {
        Label $\{\mathbf{x}, \mathbf{y}\}$ as $i$ \;
        Label $\{\mathbf{\widehat{y}}\}$ as 0 \;
    }
    Compute loss for $d_{\phi}$ by Eq. \eqref{eq:cond_cls_loss} or \eqref{eq:multi_cls_loss} .
}

\end{algorithm}

\subsubsection{Style Transformer Learning}
The training of Style Transformer is developed according to the different cases of $f_{\theta}(\mathbf{x}, \mathbf{\widehat{s}})$ where $\mathbf{s} = \mathbf{\widehat{s}}$ or $\mathbf{s} \neq \mathbf{\widehat{s}}$.

\paragraph{Self Reconstruction} For the case $\mathbf{s} = \mathbf{\widehat{s}}$ , or equivalently, the case $f_{\theta}(\mathbf{x}, \mathbf{s})$.  As we discussed before,  the input sentence $\mathbf{x}$ and the input style $\mathbf{s}$
comes from the same dataset , we can simply train our Style Transformer to reconstruct the input sentence by minimizing negative log-likelihood:
\begin{equation}
\label{eq:self_rec_loss}
\mathcal{L}_{self}(\theta) = - p_{\theta}(\mathbf{y} = \mathbf{x} | \mathbf{x}, \mathbf{s}).
\end{equation}

For the case $\mathbf{s} \neq \mathbf{\widehat{s}}$, we can't obtain direct supervision from our training set. So, we introduce two different training loss to create supervision indirectly.

\paragraph{Cycle Reconstruction} To encourage generated sentence preserving the information in the input sentence $\mathbf{x}$, we feed the generated sentence $\widehat{\mathbf{y}} = f_{\theta}(\mathbf{x},\mathbf{\widehat{s}})$ to the Style Transformer with the style of $\mathbf{x}$ and training our network to reconstruct original input sentence by minimizing negative log-likelihood:
\begin{equation}
\label{eq:cyc_rec_loss}
\mathcal{L}_{cycle}(\theta) = - p_{\theta}(\mathbf{y} = \mathbf{x} | f_{\theta}(\mathbf{x}, \mathbf{\widehat{s}}), \mathbf{s}).
\end{equation}

\paragraph{Style Controlling} If we only train our Style Transformer  to reconstruct the input sentence $\mathbf{x}$ from transferred sentence $\mathbf{\widehat{y}} = f_{\theta}(\mathbf{x},\mathbf{\widehat{s}})$, the network can only learn to copy the input to the output. To handle this degeneration problem, we further add a style controlling loss for the generated sentence. Namely, the network generated sentence $\widehat{\mathbf{y}}$ is feed into discriminator to maximize the probability of style $\mathbf{\widehat{s}}$.

For the conditional discriminator, the Style Transformer aims to minimize the negative log-likelihood of class $1$ when feed to the discriminator with the style label $\mathbf{\widehat{s}}$:
\begin{equation}
\label{eq:cond_adv_loss}
\mathcal{L}_{style}(\theta) = - p_{\phi}(\mathbf{c} = 1 | f_{\theta}(\mathbf{x}, \mathbf{\widehat{s}}), \mathbf{\widehat{s}}).
\end{equation}

And in the case of the multi-class discriminator, the Style Transformer is trained to minimize the
the negative log-likelihood of the corresponding class of style $\mathbf{\widehat{s}}$:
\begin{equation}
\label{eq:multi_adv_loss}
 \mathcal{L}_{style}(\theta) = - p_{\phi}(\mathbf{c} = \mathbf{\widehat{s}} | f_{\theta}(\mathbf{x}, \mathbf{\widehat{s}})).
\end{equation}

Combining the loss function we discussed above,
the training procedure of the Style Transformer is summarized in Algorithm \ref{alg:Transformer_network}.

\begin{algorithm}\small
\caption{Style Transformer Learning}
\label{alg:Transformer_network}
\LinesNumbered
\KwIn{Style Transformer $f_{\theta}$, discriminator $d_{\phi}$, and a dataset $\mathcal{D}_i$ with style $\mathbf{s}$}
Sample a minibatch of m sentences $\{\mathbf{x}_1, \mathbf{x}_2, ...\mathbf{x}_m\}$ from $\mathcal{D}_i$. \;
\ForEach{$\mathbf{x} \in \{\mathbf{x}_1, \mathbf{x}_2, ...\mathbf{x}_m\}$}{
    Randomly sample a style $\mathbf{\widehat{s}} (\mathbf{s} \neq \mathbf{\widehat{s}})$\;
    Use $f_{\theta}$ to generate two new sentence\\ $\mathbf{y} = f_{\theta}(\mathbf{x}, \mathbf{s})$\\
    $\mathbf{\widehat{y}} = f_{\theta}(\mathbf{x}, \mathbf{\widehat{s}})$ \;
    Compute $\mathcal{L}_{self}(\theta)$ for $\mathbf{y}$ by Eq. \eqref{eq:self_rec_loss} \;
    Compute $\mathcal{L}_{cycle}(\theta)$ for $\widehat{\mathbf{y}}$ by Eq. \eqref{eq:cyc_rec_loss} \;
    Compute $\mathcal{L}_{style}(\theta)$ for $\widehat{\mathbf{y}}$ by Eq. \eqref{eq:cond_adv_loss} or \eqref{eq:multi_adv_loss} \;
}

\end{algorithm}

\subsubsection{Summarization and Discussion}

Finally, we can construct our final training algorithm based on discriminator learning and Style Transformer learning steps. Similar to the training process of GANs \cite{goodfellow2014generative}, in each training iteration, we first perform $n_d$ steps discriminator learning to get a better discriminator, and then train our Style Transformer $n_f$ steps to improve its performance. The training process is summarized in Algorithm \ref{alg:training}.

\begin{algorithm}
\caption{Training Algorithm}\small
\label{alg:training}
\KwIn{A bunch of datasets $\{\mathcal{D}_{i}\}_{i=1}^{K}$, and each represent a different style $\mathbf{s}^{(i)}$}
Initialize the Style Transformer network $f_{\theta}$, and the discriminator network $d_{\phi}$ with random weights $\theta, \phi$ \;

\Repeat{network $f_{\theta}(\mathbf{x},
\mathbf{s})$ converges}{
      \For{$n_d$ step}{
         \ForEach{dataset $\mathcal{D}_i$}{
            Accumulate loss by Algorithm \ref{alg:discriminator}
         }
         Perform gradient decent to update $d_{\phi}$.
      }
      \For{$n_f$ step}{
        \ForEach{dataset $\mathcal{D}_i$}{
            Accumulate loss by Algorithm \ref{alg:Transformer_network}
         }
         Perform gradient decent to update $f_{\theta}$.
      }
    }

\end{algorithm}

Before finishing this section, we finally discuss a problem which we will be faced with in the training process. Because of the discrete nature of the natural language, for the generated sentence $\mathbf{\widehat{y}} = f_{\theta}(\mathbf{x},\mathbf{\widehat{s}})$, we can't directly propagate gradients from the discriminator
through the discrete samples. To handle this problem, one can use REINFORCE \cite{DBLP:journals/ml/Williams92} or the Gumbel-Softmax trick \cite{DBLP:journals/corr/KusnerH16} to estimates gradients from the discriminator. However, these two approaches are faced with high variance problem, which will make the model hard to converge. In our experiment, we also observed that the Gumbel-Softmax trick would slow down the model converging, and didn't bring much performance improvement to the model. For the reasons above, empirically, we view the softmax distribution generated by $f_{\theta}$ as a ``soft'' generated sentence and feed this distribution to the downstream network to keep the continuity of the whole training process. When this approximation is used, we also switch our decoder network from greedy decoding to continuous decoding. Which is to say, at every time step, instead of feed the token that has maximum probability in previous prediction step to the network, we feed the whole softmax distribution (Eq. \eqref{eq:softmax}) to the network. And the decoder uses this distribution to compute a weighted average embedding from embedding matrix for the input.

\section{Experiment}

\subsection{Datasets}
We evaluated and compared our approach with several state-of-the-art systems on two review datasets, Yelp Review Dataset (Yelp) and IMDb Movie Review Dataset (IMDb). The statistics of the two datasets are shown in Table \ref{tb:data}.

\noindent\textbf{Yelp Review Dataset (Yelp)} The Yelp dataset is provided by the Yelp Dataset Challenge, consisting of restaurants and business reviews with sentiment labels (negative or positive). Following previous work, we use the possessed dataset provided by \citet{li2018delete}.
Additionally, it also provides human reference sentences for the test set.

\noindent\textbf{IMDb Movie Review Dataset (IMDb)} The IMDb dataset\footnote{\url{https://github.com/fastnlp/nlp-dataset}} consists of movie reviews written by online users. To get a high quality dataset, we use the highly polar movie reviews provided by \citeauthor{imdb_dataset} \shortcite{imdb_dataset}. Based on this dataset, we construct a highly polar sentence-level style transfer dataset by the following steps: 1) fine tune a BERT \cite{DBLP:journals/corr/abs-1810-04805} classifier on original training set, which achieves $95\%$ accuracy on test set; 2) split each review in the original dataset into several sentences; 3) filter out sentences with confidence threshold below 0.9 by our fine-tuned BERT classifier; 4) remove sentences with uncommon words.
Finally, this dataset contains 366K, 4k, 2k sentences for training, validation, and testing, respectively.

\begin{table}[t!]\small
  \centering
  \begin{tabular}{ccccc}
    \toprule
    \multirow{2}{*}{Dataset} & \multicolumn{2}{c}{Yelp} &  \multicolumn{2}{c}{IMDb}                  \\
    \cmidrule(lr){2-3}\cmidrule(lr){4-5}
         & Positive    & Negative  & Positive    & Negative  \\
    \midrule
    Train  & 266,041  & 177,218 & 178,869 & 187,597     \\
    Dev      & 2,000 & 2,000 & 2,000  & 2,000    \\
    Test      &  500 & 500 & 1,000 &  1,000   \\
    \cmidrule(lr){2-3}\cmidrule(lr){4-5}
    Avg. Len.      &  \multicolumn{2}{c}{8.9} &  \multicolumn{2}{c}{18.5}   \\
    \bottomrule
  \end{tabular}\caption{Datasets statistic.}\label{tb:data}
\end{table}

\begin{table*}[ht!]\small
  \centering
  \begin{tabular}{lccccccc}
    \toprule
    \multirow{2}{*}{Model}& \multicolumn{4}{c}{Yelp} &  \multicolumn{3}{c}{IMDb} \\
    \cmidrule(lr){2-5} \cmidrule(lr){6-8}
         & ACC   & \textit{ref}-BLEU & \textit{self}-BLEU & PPL & ACC    & \textit{self}-BLEU  & PPL  \\
   \midrule
    Input Copy                              & 3.8  & 23 & 100 & 41 & 5.1  & 100 & 58  \\

    \midrule
    RetrieveOnly \cite{li2018delete}        & 92.6  & 0.4 & 0.7 & \textbf{7} & N/A & N/A & N/A  \\
    TemplateBased \cite{li2018delete}       & 84.3  & 13.7 & 44.1 & 117  & N/A & N/A & N/A \\
    DeleteOnly \cite{li2018delete}          & 85.7  & 9.7 & 28.6 & 72 & N/A & N/A & N/A  \\
    DeleteAndRetrieve \cite{li2018delete}   & 87.7    & 10.4 & 29.1 & 60 & 58.8 & 55.4 & \textbf{57}  \\
    \midrule

    ControlledGen \cite{hu2017toward}       & 88.8  & 14.3 & 45.7 & 219 & 94.1  & 62.1 & 143 \\
    CrossAlignment \cite{shen2017style}     & 76.3  & 4.3 & 13.2 & 53 & N/A & N/A & N/A \\
    MultiDecoder \cite{Fu18}                & 49.8 & 9.2 & 37.9 & 90  & N/A & N/A & N/A  \\
    CycleRL\cite{xu2018unpaired}            & 88.0    & 2.8 & 7.2 & 107  & \textbf{97.8}  & 4.9 & 177 \\
    \midrule

    Ours (Conditional)                      & \textbf{93.7}  & 17.1 & 45.3 & 90 & 86.6  & 66.2 & 107  \\
    Ours (Multi-Class)                      & 87.7  & \textbf{20.3} & \textbf{54.9} & 73 & 80.3  & \textbf{70.5} & 105  \\

    \bottomrule
  \end{tabular}
  \caption{Automatic evaluation results on Yelp and IMDb datset}
  \label{tab:auto_result}
\end{table*}

\subsection{Evaluation}
A goal transferred sentence should be a fluent, content-complete one with target style.
To evaluate the performance of the different model, following previous works, we compared three different dimensions of generated samples: 1) Style control, 2) Content preservation and 3) Fluency.

\subsubsection{Automatic Evaluation}

\noindent\textbf{Style Control} We measure style control automatically by evaluating the target sentiment accuracy of transferred sentences. For an accurate evaluation of style control, we trained two sentiment classifiers on the training set of Yelp and IMDb using fastText \cite{joulin2017bag}.

\noindent\textbf{Content Preservation} To measure content preservation, we calculate the BLEU score  \cite{papineni2002bleu} between the transferred sentence and its source input using NLTK.  A higher BLEU score indicates the transferred sentence can achieve better content preservation by retaining more words from the source sentence. If a human reference is available, we will calculate the BLEU score between the transferred sentence and corresponding reference as well. Two BLEU score metrics are referred to as \textit{self}-BLEU and \textit{ref}-BLEU respectively.

\noindent\textbf{Fluency} Fluency is measured by the perplexity of the transferred sentence, and we trained a 5-gram language model on the training set of two datasets using KenLM \cite{heafield2011kenlm}.

\subsubsection{Human Evaluation}
Due to the lack of parallel data in style transfer area, automatic metrics are insufficient to evaluate the quality of the transferred sentence. Therefore we also conduct human evaluation experiments on two datasets.

We randomly select 100 source sentences (50 for each sentiment) from each test set for human evaluation. For each review, one source input and three anonymous transferred samples are shown to a reviewer. And the reviewer is asked to choose the best sentence for style control, content preservation, and fluency respectively.
\begin{itemize}
  \item Which sentence has the most opposite sentiment toward the source sentence?
  \item Which sentence retains most content from the source sentence?
  \item Which sentence is the most fluent one?
\end{itemize}
To avoid interference from similar or same generated sentences, "no preference." is also an option answer to these questions.

\subsection{Training Details}
In all of the experiment, for the encoder, decoder, and discriminator, we all use 4-layer Transformer with four attention heads in each layer. The hidden size, embedding size, and positional encoding size in Transformer are all 256 dimensions. Another embedding matrix with 256 hidden units is used to represent different style, which is feed into encoder as an extra token of the input sentence. And the positional encoding isn't used for the style token. For the discriminator, similar to  \citet{gpt} and \citet{DBLP:journals/corr/abs-1810-04805}, we further add a \texttt{<cls>} token to the input, and the output vector of the corresponding position is feed into a softmax classifier which represents the output of discriminator.

In the experiment, we also found that preforming random word dropout for the input sentence when computing the self reconstruction loss (Eq. \eqref{eq:self_rec_loss}) can help model more easily to converge to a reasonable performance. On the other hand, by adding a temperature parameter to the softmax layer (Eq.  \eqref{eq:softmax}) and using a sophisticated temperature decay schedule can also help the model to get a better result in some case.  

\subsection{Experimental Results}
Results using \textbf{automatic metrics} are presented in Table \ref{tab:auto_result}. Comparing to previous approaches, our models achieve competitive performance overall and get better content preservation at all of two datasets. Our conditional model can achieve a better style controlling compared to the multi-class model.
Both our models are able to generate sentences with relatively low perplexity. For those previous models performing the best on a single metric, an obvious drawback can always be found on another metric.

For the \textbf{human evaluation},
we choose two of the most well-performed models according to the automatic evaluation results as competitors: {DeleteAndRetrieve} (DAR) \cite{li2018delete} and {Controlled Generation} (CtrlGen) \cite{hu2017toward}. And the generated outputs from multi-class discriminator model is used as our final model.
We have performed over 400 human evaluation reviews. Results are presented in Table \ref{tab:human_result}. The human evaluation results are mainly conformed with our automatic evaluation results. And it also shows that our models are better in content preservation, compared to two competitor model.

Finally, to better understand the characteristic of different models, we sampled several output sentences from the Yelp dataset, which are shown in Table \ref{tb:yelp_samples}.

\begin{table}[t!]\small\setlength{\tabcolsep}{2pt}
  \centering
  \begin{tabular}{lcccccc}
    \toprule
    \multirow{2}{*}{Model}& \multicolumn{3}{c}{Yelp} &  \multicolumn{3}{c}{IMDb}                  \\

    \cmidrule(lr){2-4} \cmidrule(lr){5-7}
    & Style    & Content  & Fluency   & Style    & Content  & Fluency\\
    \midrule
    CtrlGen         & 16.8  & 23.6 & 17.7  & \textbf{30.0}  & 19.5 & 22.0   \\
    DAR         & 13.6  & 15.5 & 21.4  & 21.0  & 27.0 & 25.0   \\
    Ours    & \textbf{48.6}  & \textbf{36.8} & \textbf{41.4}  & 29.5  & \textbf{35.0} & \textbf{31.5}  \\
    \midrule
    No Preference    & 20.9  & 24.1 & 19.5  & 19.5  & 18.5 & 21.5  \\
    \bottomrule
  \end{tabular}\caption{Human evaluation results on two datasets. Each cell indicates the proportion of being preferred.}
  \label{tab:human_result}
\end{table}

\newcommand{\good}[1]{{\color{red}{#1}}}
\newcommand{\bad}[1]{{\color{blue}{#1}}}
\newcommand{\grammar}[1]{{\color{brown}{#1}}}

\begin{table*}[t!]\small\setlength{\tabcolsep}{2pt}
  \centering
  \begin{tabular}{rl}
    \toprule
    \multicolumn{2}{c}{\textbf{negative to positive}}\\
    \midrule
    \textbf{Input}& the food 's ok , the service is among the worst i have encountered . \\
    \textbf{DAR}  & the food 's \bad{ok} , the service is \grammar{among great} and service \grammar{among} . \\
    \textbf{CtrlGen} & the food 's \bad{ok} , the service is among the \grammar{randy} i have encountered . \\
    \textbf{Ours} & the food 's \good{delicious} , the service is among \good{the best} i have encountered . \\
    \textbf{Human}& the food is good , and the service is one of the best i 've ever encountered . \\
    \midrule
    \textbf{Input}& this is the worst walmart neighborhood market out of any of them . \\
    \textbf{DAR}& walmart market is one of my favorite places \grammar{in any neighborhood out of them} . \\
    \textbf{CtrlGen}& \grammar{fantastic} is the randy go neighborhood market out of any of them . \\
    \textbf{Ours} &this is the \good{best} walmart neighborhood market out of any of them . \\
    \textbf{Human} &this is the best walmart out of all of them . \\
     \midrule
     \textbf{Input}& always rude in their tone and always have shitty customer service ! \\
     \textbf{DAR}& i always enjoy going in \grammar{always} their \bad{kristen} and always have \bad{shitty} customer service ! \\
     \textbf{CtrlGen}& always \good{good} in their tone and always have \bad{shitty} customer service ! \\
     \textbf{Ours}& always \good{nice} in their tone and always have \grammar{provides} customer service ! \\
     \textbf{Human}& such nice customer service , they listen to anyones concerns and assist them with it .\\
    \midrule
    \multicolumn{2}{c}{\textbf{positive to negative}}\\
    \midrule
    \textbf{Input}& everything is fresh and so delicious ! \\
    \textbf{DAR}& small impression was ok , \bad{but lacking i have piss stuffing night} . \\
    \textbf{CtrlGen}& everything is disgrace and so bland ! \\
    \textbf{Ours}& everything is \good{overcooked} and so \good{cold} ! \\
    \textbf{Human}& everything was so stale .\\
     \midrule
     \textbf{Input}& these two women are professionals . \\
     \textbf{DAR}& these two \bad{scam women} are professionals . \\
     \textbf{CtrlGen}& \bad{shame two women} are unimpressive . \\
     \textbf{Ours} &these two women are \good{amateur} . \\
     \textbf{Human} &these two women are not professionals .\\
    \midrule
    \textbf{Input} &fantastic place to see a show as every seat is a great seat ! \\
    \textbf{DAR} &\bad{there is no reason} to see a show as every \grammar{seat seat} ! \\
    \textbf{CtrlGen}& unsafe place to \grammar{embarrassing lazy run} as every seat is \grammar{lazy disappointment} seat ! \\
    \textbf{Ours} &\good{disgusting} place to see a show as every seat is a \good{terrible} seat ! \\
    \textbf{Human} &terrible place to see a show as every seat is a horrible seat !\\
    \bottomrule
  \end{tabular}\caption{Case study from Yelp dataset. The \good{red} words indicate good transfer; the \bad{blue} words indicate bad transfer; the \grammar{brown} words indicate grammar error.}
  \label{tb:yelp_samples}
\end{table*}

\subsection{Ablation Study}

\begin{table}[t!]\small\setlength{\tabcolsep}{2pt}
  \centering
  \begin{tabular}{l*6{c}}
    \toprule
    & \multicolumn{3}{c}{Conditional} &  \multicolumn{3}{c}{Multi-class}                  \\
    \cmidrule(lr){2-4}\cmidrule(lr){5-7}
    Model     & ACC   & BLEU  & PPL & ACC   & BLEU  & PPL\\
    \midrule
    Style Transformer     & 93.6  & 17.1 & 78  & 87.6  & 20.3 & 50 \\
    \midrule
    - self reconstruction       & 50.0  & 0 & N/A  & 20.7  & 0 & N/A \\
    - cycle reconstruction       & 94.2  & 8.6 & 56 & 93.2 & 8.7 & 40   \\
    - discriminator       & 3.3  & 22.9 & 11 & 3.3  & 22.9 & 11 \\
    \midrule
    - real sample       & 89.7  & 17.4 & 75  & 83.8  & 19.4 & 55\\
    - generated sample      & 46.3  & 21.6 & 34 & 35.6  & 22.0 & 33  \\
    \bottomrule
  \end{tabular}
  \caption{Model ablation study results on Yelp dataset}
  \label{tab:ablation_result}
\end{table}

To study the impact of different components on overall performance, we further did an ablation study for our model on Yelp dataset, and results are reported in Table \ref{tab:ablation_result}. 


For better understanding the role of different loss functions, we disable each loss function by turns and retrain our model with the same setting for the rest of hyperparameters. After we disable self-reconstruction loss (Eq. \eqref{eq:self_rec_loss}), our model failed to learn a meaningful output and only learned to generate a single word for any combination of input sentence and style.
However, when we don't use cycle reconstruction loss (Eq. \eqref{eq:cyc_rec_loss}), it's also possible to train the model successfully, and both of two models
converge to reasonable performance. And comparing to the full model, there is a small improvement in style accuracy, but a significant drop in BLEU score.
As our expected, the cycle reconstruction loss is able to encourage the model to preserve the information from the input sentence.
At last, when the discriminator loss (Eq. \eqref{eq:cond_adv_loss} and \eqref{eq:multi_adv_loss}) is not used, the model quickly degenerates to a model which is only copying the input sentence to output without any style modification. This behaviour also conforms with our intuition. If the model is only asked to minimize the self-reconstruction loss and cycle reconstruction loss, directly copying input is one of the optimum solutions which is the easiest to achieve. In summary, each of these loss plays an important role in the Style Transformer training stage: 1) the self-reconstruction loss guides the model to generate readable natural language sentence. 2) the cycle reconstruction loss encourages the model to preserve the information in the source sentence. 3) the discriminator provides style supervision to help the model control the style of generated sentences.

Another group of study is focused on the different type of samples used in the discriminator training step. In Algorithm \ref{alg:discriminator}, we used a mixture of real sentence $\mathbf{x}$ and generated sentence $\mathbf{y}$ as the positive training samples for the discriminator.
By contrast, in the ablation study, we trained our model with only one of them. As the result shows, the generated sentence is the key component in discriminator training. When we remove the real sentence from the training data of discriminator, our model can also achieve a competitive result as the full model with only a small performance drop. However,
if we only use the real sentence
the model will lose a significant part of the ability to control the style of the generated sentence, and thus yields a bad performance in style accuracy. However, the model can still perform a style control far better than the input copy model discussed in the previous part.
For the reasons above, we used a mixture of real sample and generated sample in our final version.


\section{Conclusions and Future Work}

In this paper, we proposed the Style Transformer with a novel training algorithm for text style transfer task. 
Experimental results on two text style transfer datasets have shown that our model achieved a competitive or better performance compared to previous state-of-the-art approaches. Especially, because our proposed approach doesn't assume a disentangled latent representation for manipulating the sentence style, our model can get better content preservation on both of two datasets.

In the future, we are planning to adapt our Style Transformer to the multiple-attribute setting like \citet{lample2018multipleattribute}. On the other hand, the back-translation technique developed in \citet{lample2018multipleattribute} can also be adapted to the training process of Style Transformer. How to combine the back-translation with our training algorithm is also a good research direction that is worth to explore.

\section*{Acknowledgment}

We would like to thank the anonymous reviewers for their valuable comments. The research work is supported by National Natural Science Foundation of China (No. 61751201 and 61672162),
Shanghai Municipal Science and Technology Commission (16JC1420401 and 17JC1404100),
Shanghai Municipal Science and Technology Major Project(No.2018SHZDZX01)and ZJLab.

\bibliography{acl2019}
\bibliographystyle{acl_natbib}

\end{document}